\documentclass[10pt, a4paper, twocolumn]{article} % twocolumn nativo

\usepackage[utf8]{inputenc}
\usepackage[T1]{fontenc}
\usepackage[english]{babel}

\usepackage{tgheros} % Helvetica

\usepackage[varqu]{zi4}
\usepackage{microtype}

\usepackage{titlesec}
\titlespacing*{\section}{0pt}{1.5ex plus 1ex minus .2ex}{0.1em}
\titlespacing*{\subsection}{0pt}{1.5ex plus 1ex minus .2ex}{0.1em}
\titlespacing*{\subsubsection}{0pt}{1.5ex plus 1ex minus .2ex}{0.1em}

\usepackage[table]{xcolor}
\usepackage{booktabs}
\usepackage{float}
\usepackage{graphicx}
\usepackage[labelfont={bf,small}, textfont={small}]{caption}
\usepackage{subcaption}

\definecolor{acqua_celeste}{RGB}{0, 160, 175}

\usepackage{amsmath}
\usepackage{amssymb}
\usepackage{mathtools}
\usepackage{amsthm}
\usepackage{bbold}
\usepackage{pifont}
\usepackage{multirow}
\usepackage{arydshln}
\usepackage{array}
\usepackage{makecell}
\usepackage{wrapfig}
\usepackage{tabularx}

\definecolor{pgreen}{rgb}{0.13, 0.55, 0.13}
\definecolor{pred}{rgb}{0.8, 0.13, 0.13}
\definecolor{diversity}{HTML}{D4E1F5}
\definecolor{recognizability}{HTML}{FFE6CC}

\newcommand{\xmark}{\ding{55}}
\newcommand{\cmark}{\ding{51}}

\newcolumntype{Y}{>{\centering\arraybackslash}X}

\theoremstyle{plain}

\theoremstyle{definition}

\theoremstyle{remark}

\usepackage[left=1.5cm, right=1.5cm, top=1.5cm, bottom=2cm]{geometry}
\usepackage[numbers, sort&compress]{natbib}
\usepackage{hyperref}
\usepackage[capitalize,noabbrev]{cleveref}

\hypersetup{
    colorlinks=true,
    linkcolor=black,
    urlcolor=acqua_celeste,
    citecolor=acqua_celeste
}

\usepackage{fancyhdr}
\usepackage{authblk}

\title{\huge\bfseries\vspace{-1em} A Filtered Mixture-of-Generators for Fully Synthetic Survival Training}

\author[1]{Niccol\`o Maria Rizzi}
\author[1]{Eugenio Lomurno\textsuperscript{*}}
\author[1]{Alberto Archetti}
\author[1]{Matteo Matteucci}

\affil[1]{Politecnico di Milano, Milan, Italy}

\date{}

\begin{document}

% Blocco per Titolo e Abstract a tutta larghezza
\twocolumn[
  \begin{@twocolumnfalse}
    \maketitle
    \vspace{-2em}

    \begin{abstract}
        \setlength{\parindent}{0pt}
        \setlength{\parskip}{4pt}
        \itshape

        \noindent\textnormal{\textbf{Objective:}} Survival analysis is a statistical framework for time-to-event modelling in a wide range of critical domains. In clinical settings, training data are particularly costly to assemble, since events accrue over years of follow-up, cohort sizes remain small, and privacy regulations restrict sharing across institutions. Tabular generative models offer, in principle, both augmentation and privacy-preserving cohort sharing, but are themselves data-hungry: on the small cohorts typical of survival analysis, a single generator rarely characterizes the population well enough for downstream models trained on its output to match real-data performance. We aim to make fully synthetic training a viable substitute for real-data training in this regime.

        \noindent\textnormal{\textbf{Methods:}} We propose FoGS (Filtered Mixture-of-Generators for Survival analysis), a two-level pipeline that reframes synthetic-data construction as sample selection rather than sample generation. A candidate pool is drawn from four architecturally distinct tabular generators, and each sample is scored by an ensemble of seven survival models trained on real data, using proper scoring rules as a per-sample plausibility proxy. An outer loop optimizes a selection policy---generator quotas, scorer weights, a random complement, and stratified balancing on event time and censoring---against held-out downstream performance, while an inner loop tunes the downstream survival model (XGBoost-Cox). We evaluate FoGS on 16 public datasets under train-on-synthetic, test-on-real, reporting C-index and IBS on a 0--100 scale.

        \noindent\textnormal{\textbf{Results:}} FoGS yields mean improvements of $+2.17$ in C-index and $+0.67$ in IBS, improving both metrics on 9 of 16 datasets and at least one on 13 (one-sided Wilcoxon $p=0.039$ and $p=0.035$). It matches or exceeds real-data training on most cohorts, with no significant change in nearest-neighbour privacy margin relative to unfiltered sampling.

        \noindent\textnormal{\textbf{Conclusion:}} Sample filtering over a heterogeneous generator pool is a viable substitute for real-data training in privacy-restricted clinical settings.
    \end{abstract}

    \vspace{0.5em}

    \vspace{1em}
    \hrule height 1pt
    \vspace{2em}
  \end{@twocolumnfalse}
]

% --- FOOTNOTE (corresponding) ---
{
  \renewcommand{\thefootnote}{\fnsymbol{footnote}}
  \footnotetext[1]{Corresponding author: \texttt{eugenio.lomurno@polimi.it}.}
}
% ------------------------------------------

\section{Introduction}
\label{sec:introduction}

Survival models guide clinical decision-making across oncology, cardiology, transplantation, and other specialties, supporting treatment stratification, follow-up scheduling, and clinical-trial design \cite{wang2019machine, klein2003survival}. Their inputs are right-censored time-to-event data: tuples of covariates, observed time, and an event indicator, with the event time only partially known for subjects still under observation at study end \cite{cox1972regression, kaplan1958nonparametric}. Training survival models is constrained by structural properties of the data itself: clinically meaningful events accrue only over years of follow-up, which keeps cohort sizes small, and privacy regulations restrict sharing across institutions. These constraints make the curation of a sufficiently large training cohort the slowest and most expensive step in deploying a survival model in practice.

Synthetic tabular generation has been proposed as a remedy, with model fidelity steadily improving across the heterogeneous architectures developed for structured data. In principle, a high-quality synthetic cohort serves two complementary roles: augmenting scarce real-data training sets to improve downstream model robustness, and enabling cohort sharing across institutions without disclosing patient-level records \cite{yoon2020anonymization}. The viability of either role depends on whether synthetic data, when substituted for real training data, preserves the downstream task performance achievable on real data.

Yet tabular generative approaches remain imperfect even on large-scale datasets, and the relatively small cohorts typical of survival data amplify the problem. With few examples, generators cannot learn a faithful representation of the underlying population distribution, and the resulting synthetic data fails to cover the regions of the data manifold on which downstream models depend. Train-on-synthetic, test-on-real (TSTR) evaluation with single generative models consistently underperforms training on real data, both in survival generation \cite{norcliffe2023survivalgan, qian2023synthcity} and in broader tabular and image domains \cite{lomurno2026inference, resmini2025your}. Closing this gap requires reframing the construction of a synthetic survival training set: not as the output of a single generator, but as a selection problem optimized against the downstream task it is meant to serve.

We introduce FoGS (Filtered Mixture-of-Generators for Survival analysis): a two-level pipeline that reframes synthetic-data construction as a sample-selection problem rather than a sample-generation problem. A heterogeneous pool of synthetic samples is drawn from four architecturally distinct tabular generators, and each sample is scored by an ensemble of survival models trained on real data. An outer optimization loop tunes the selection policy (generator quotas, scorer weights, a random complement, and stratified balancing) against held-out downstream performance, while an inner loop reproduces the standard survival-modeling pipeline a practitioner would apply to the resulting synthetic dataset (Fig.~\ref{fig:pipeline}). This nested design ensures the outer-loop signal reflects the downstream utility actually obtainable in standard practice. On 16 public survival datasets, FoGS improves both downstream metrics on 9 cohorts and at least one on 13, matching or exceeding real-data training on most of the collection and supporting synthetic data as a substitute for real cohorts in privacy-restricted clinical settings.

This paper makes three contributions. First, we introduce FoGS, to our knowledge the first pipeline that optimizes synthetic-data construction for survival analysis, recasting the problem as sample selection over a heterogeneous generator pool, driven by a per-sample plausibility signal. Second, we benchmark FoGS across 16 public survival datasets, demonstrating improvements in downstream performance on the majority of cohorts. Third, we identify two structural phenomena that govern synthetic-data selection in this regime: a trade-off between per-sample plausibility and population coverage, and a dataset-dependence under which no single strategy prevails across cohorts.

\begin{table}[h]
\centering
\caption{Statement of significance}
\label{tab:significance}
\begin{tabularx}{\columnwidth}{@{}>{\raggedright\arraybackslash\bfseries}p{0.28\columnwidth}X@{}}
\toprule
Problem or issue & Survival models for clinical decision-making require large training cohorts, but events accrue over years, cohorts stay small, and privacy rules restrict cross-institution data sharing. \\
\addlinespace
What is already known & Tabular generative models can produce synthetic cohorts, but a single generator trained on small survival data yields synthetic sets whose train-on-synthetic, test-on-real utility falls below real-data training. \\
\addlinespace
What this paper adds & FoGS reframes synthetic-data construction as sample selection over a pool of four heterogeneous generators, scoring each sample with real-data survival models and tuning the selection policy against downstream utility. It matches or exceeds real-data training on most of 16 public datasets without degrading the nearest-neighbour privacy margin. \\
\addlinespace
Who would benefit & Clinical researchers and data scientists building survival models from scarce, privacy-restricted cohorts, and institutions seeking to share synthetic cohorts safely. \\
\bottomrule
\end{tabularx}
\end{table}

\section{Related Work}
\label{sec:related-work}

\subsection{Synthetic Tabular Generation}
\label{subsec:rw-tabular-generation}

Tabular data generation has been approached through several distinct directions: explicit factorization via Bayesian networks \cite{Ankan2024}, adversarial training in GAN-family architectures including ADS-GAN \cite{yoon2020anonymization} and CTGAN \cite{xu2019modeling}, and latent-variable modelling via variational autoencoders \cite{kingma2013auto, akrami2020robust}. More recently, invertible transformations via normalizing flows \cite{durkan2019neural}, denoising diffusion \cite{sohl2015deep, kotelnikov2023tabddpm}, and non-parametric density estimation via adversarial random forests \cite{watson2023adversarial} have extended this landscape. Quality has improved steadily across paradigms, yet no single architecture dominates across tabular structures, with each capturing different aspects of feature dependence. FoGS exploits this fragmentation by pooling samples from four generators spanning these families: a Bayesian network \cite{Ankan2024, tsamardinos2006max}, adversarial random forests \cite{watson2023adversarial}, TabDDPM \cite{kotelnikov2023tabddpm}, and a survival-aware CTGAN variant \cite{norcliffe2023survivalgan}. FoGS leverages their complementary inductive biases rather than committing to any one.

\subsection{Synthetic Survival Generation and Utility Evaluation}
\label{subsec:rw-survival-and-utility}

Survival generation must reproduce both covariates and the censored time-to-event signal, and existing approaches handle censoring in two distinct ways. Dedicated architectures are primarily GAN-based, with SurvivalGAN as the canonical instance \cite{norcliffe2023survivalgan}. These models handle censoring explicitly: covariates are generated by a GAN component and event times are derived through a separate survival head. General-purpose tabular generators have instead been adapted to survival by jointly estimating the event indicator and observed time alongside the covariates, absorbing censoring into the joint distribution rather than modelling it explicitly \cite{qian2023synthcity}. TSTR utility under either strategy is commonly reported below real-data baselines, and evaluation practice treats utility as a post-hoc quality-control measurement \cite{qian2023synthcity}. FoGS departs from both lines: rather than introducing a new survival-specific generator or treating utility as a post-hoc check, it composes generators of both kinds and reframes synthetic dataset construction as policy-tuned selection over their pooled output. TSTR utility is treated as the outer optimization objective.

\section{Method}
\label{sec:method}

\begin{figure*}[t]
    \centering
    \includegraphics[width=\linewidth]{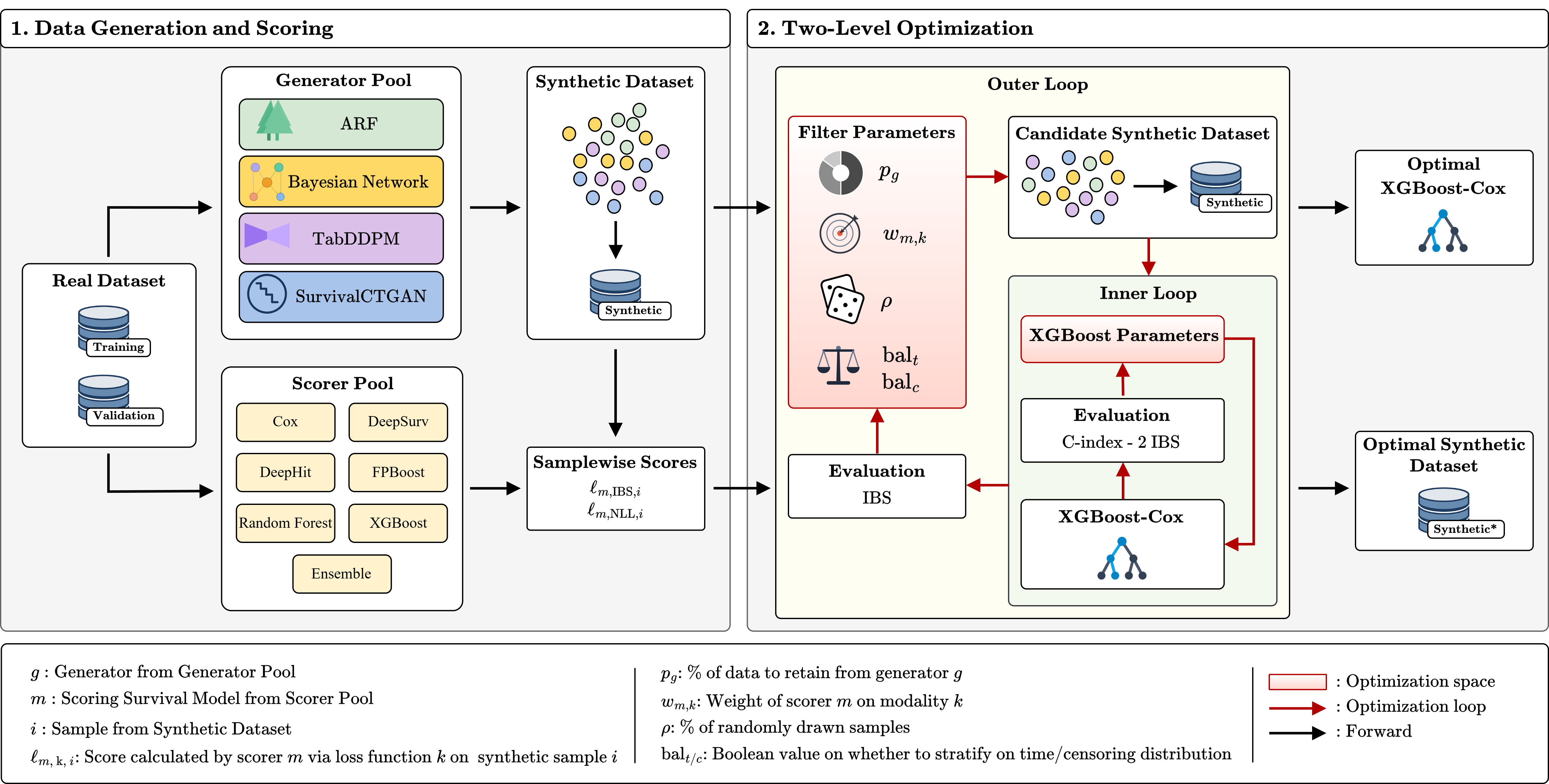}
    \caption{Overview of the FoGS pipeline. Stage~1 (left) draws a candidate pool from the four generators and scores every sample with the survival-model ensemble trained on the real cohort. Stage~2 (right) filters the pool under the selection policy and trains the downstream model, with the two-level optimization driven by feedback on the real validation split.}
    \label{fig:pipeline}
\end{figure*}

FoGS constructs a high-utility synthetic survival training set by generating and selecting the samples that best serve a downstream survival model. We cast this as a two-level optimization (Fig.~\ref{fig:pipeline}): an outer level searches over selection policies to maximize downstream utility on real data, and an inner level trains a representative downstream model to guide the outer search.

\subsection{Synthetic Data and Generator Pool}

Let $\mathcal{D}_r = \{(\mathbf{x}_i, t_i, \delta_i)\}_{i=1}^{N_r}$ denote the real training split, with $\mathbf{x}_i$ features, $t_i$ observed time, and $\delta_i$ event indicator. $\mathcal{D}_r$ is held out from the validation and test splits, used respectively for hyperparameter tuning and final out-of-sample evaluation. FoGS starts from a generator pool $\mathcal{G}$, where each generator $g \in \mathcal{G}$ is trained on $\mathcal{D}_r$ and defines a sampling distribution over synthetic instances. From each $g$ we draw a candidate pool $\mathcal{D}_g$ of size $N_g = \lceil\kappa\,N_r\rceil$, where $\kappa \geq 1$ controls the size of the candidate pool relative to the real cohort. The union $\mathcal{D}_{\mathrm{synth}} = \bigcup_{g \in \mathcal{G}} \mathcal{D}_g$ is the candidate set. In our experiments, the generator pool is instantiated with four generators spanning the principal paradigms of tabular generation---explicit factorization, adversarial training, denoising diffusion, and non-parametric density estimation---so that $\mathcal{D}_{\mathrm{synth}}$ is not tied to the inductive bias of any single generator. The generators included in $\mathcal{G}$ are the following:
\begin{itemize}
    \item \emph{ARF} \cite{watson2023adversarial}: non-parametric density estimation via iterative adversarial training of a random-forest classifier, appropriate for mixed continuous and categorical features.
    \item \emph{Bayesian Network} \cite{Ankan2024, tsamardinos2006max}: explicit factorization with hill-climbing structure learning, providing an interpretable factorized representation with native support for discrete variables.
    \item \emph{TabDDPM} \cite{kotelnikov2023tabddpm}: denoising diffusion with multinomial and Gaussian forward processes for categorical and numerical features respectively.
    \item \emph{SurvivalCTGAN}: CTGAN \cite{xu2019modeling} adapted to censored survival data following SurvivalGAN methodology \cite{norcliffe2023survivalgan}, with features generated by the GAN component and event times derived through a survival head.
\end{itemize}

\subsection{Per-Sample Plausibility Scoring}
\label{subsec:method-scoring}

After generating $\mathcal{D}_{\mathrm{synth}}$, FoGS uses a set of survival models, defined as scorers, to assess the plausibility of each synthetic sample. Let $\mathcal{M}$ be the scorer pool. Each scorer $m \in \mathcal{M}$, trained on $\mathcal{D}_r$, assigns to the $i$-th synthetic sample two per-sample losses, the integrated Brier score (IBS) $\ell_{m,\mathrm{IBS},i}$ and the negative log-likelihood (NLL) $\ell_{m,\mathrm{NLL},i}$. Both IBS and NLL are proper scoring rules \cite{glenn1950verification, gneiting2007strictly}, whose expected per-sample value is minimized exactly at the true conditional distribution. A sample to which a real-data-trained scorer assigns low loss is therefore plausible with respect to the real population.

In our experiments, the scorer pool is instantiated with seven survival models---Cox \cite{cox1972regression}, DeepSurv \cite{katzman2018deepsurv}, DeepHit \cite{lee2018deephit}, RSF \cite{ishwaran2008random}, FPBoost \cite{archetti2024fpboost}, XGBoost-Cox \cite{chen2016xgboost}, and a weighted ensemble of the previous six---covering linear, tree-based, and neural survival methods. Training and hyperparameter tuning are performed once per dataset.

\subsection{Selection Policy}

The filtered synthetic dataset is
\begin{equation}
    \mathcal{D}_s^{*} \;=\; \Phi(\mathcal{D}_{\mathrm{synth}};\, \pi),
    \label{eq:framework}
\end{equation}
where $\Phi$ is a selection operator defined by a policy vector
\begin{equation}
    \pi \;=\; \big(\{p_g\},\; \{w_{m,k}\},\; \rho,\; \mathrm{bal}_t,\; \mathrm{bal}_c\big).
    \label{eq:policy}
\end{equation}
Here, $\pi$ collects the sample-selection parameters. The quotas $p_g \in [0,1]$, for each $g \in \mathcal{G}$, set the number of samples drawn from each candidate pool $\mathcal{D}_g$ as $\lceil p_g\,N_g\rceil$. Each sample is assigned an aggregated composite loss
\begin{equation}
    \mathcal{L}_i \;=\; \frac{\sum_{m,k}\, w_{m,k}\, \ell_{m,k,i}}{\sum_{m,k}\, w_{m,k}},
    \label{eq:perSampleLoss}
\end{equation}
where $m \in \mathcal{M}$ are the scorer survival models, $\ell_{m,k,i}$ is the per-sample loss of type $k\in \{\mathrm{IBS}, \mathrm{NLL}\}$, and $w_{m,k} \in [0,1]$ the aggregation weights from $\pi$.

Within each per-generator draw, a fraction $(1-\rho)$ is selected deterministically among the most plausible samples, and the remaining fraction is drawn uniformly at random from the leftover pool. Thus, the selected set is
\begin{equation}
    \mathcal{D}_s^{*} \;=\; \bigcup_{g \in \mathcal{G}} ( \mathcal{T}_g \cup \mathcal{R}_g ),
    \label{eq:selection}
\end{equation}
where $\mathcal{T}_g$ is selected so that $\lvert\mathcal{T}_g\rvert = \lceil(1-\rho)\,p_g\,N_g\rceil$, and $\mathcal{R}_g$ is sampled uniformly from $\mathcal{D}_g \setminus \mathcal{T}_g$, with $\lvert\mathcal{R}_g\rvert = \lfloor\rho\,p_g\,N_g\rfloor$. The fraction $\rho$ thus tunes how strongly selection relies on plausibility versus uniform coverage.

Finally, the flags $\mathrm{bal}_t, \mathrm{bal}_c \in \{0, 1\}$ optionally constrain the selection to be stratified: $\mathrm{bal}_t$ enables stratification with respect to the event-time distribution of $\mathcal{D}_r$, while $\mathrm{bal}_c$ stratifies with respect to the censoring rate.

\subsection{Two-Level Optimization}
\label{subsec:method-optimization}

At a high level, FoGS optimizes the selection policy $\pi$ with two nested optimizations. The inner optimization mimics a practitioner training a representative survival model and provides feedback for the outer policy search; at convergence, $\pi$ is the selection policy under which a generic survival model attains its best validation performance.

In particular, the inner loop trains a downstream survival model $\sigma_\theta$ on the selected set $\mathcal{D}_s^{*}$ and tunes its hyperparameters $\theta$. In our experiments, the downstream model is instantiated as XGBoost-Cox \cite{chen2016xgboost}, a strong and widely-used tabular survival model that serves as an efficient, representative proxy for a standard survival pipeline. Although XGBoost-Cox also appears in the scorer pool $\mathcal{M}$, the two roles are distinct: scorers are trained on real data to rank synthetic candidates, whereas the downstream model is trained on the selected synthetic set. The inner loop returns
\begin{equation}
    \theta^{*} \;=\; \arg\max_{\theta} \big( \mathrm{C}_{\mathrm{val}}(\sigma_\theta) \;-\; 2\,\mathrm{IBS}_{\mathrm{val}}(\sigma_\theta) \big),
    \label{eq:innerObj}
\end{equation}
where $\mathrm{C}_{\mathrm{val}}$ and $\mathrm{IBS}_{\mathrm{val}}$ are the concordance index and IBS of $\sigma_\theta$ computed on the real validation split. The factor of 2 normalizes the two metrics to comparable scales.

The outer loop, instead, optimizes the selection policy
\begin{equation}
    \pi^{*} \;=\; \arg\min_{\pi} \;\mathrm{IBS}_{\mathrm{val}}\big( \sigma_{\theta^{*}} \big),
    \label{eq:outerObj}
\end{equation}
where, for each $\pi$, the candidate set is filtered into $\mathcal{D}_s^{*} = \Phi(\mathcal{D}_{\mathrm{synth}};\, \pi)$, the inner loop \eqref{eq:innerObj} yields $\theta^{*}$ on $\mathcal{D}_s^{*}$, and the resulting model $\sigma_{\theta^{*}}$ is evaluated on the real validation split. We adopt $\mathrm{IBS}_{\mathrm{val}}$ as the outer objective, in preference to $\mathrm{C}_{\mathrm{val}}$ alone or the composite $\mathrm{C}_{\mathrm{val}} - 2\,\mathrm{IBS}_{\mathrm{val}}$. This choice is verified empirically on three representative datasets in Appendix~\ref{app:outer-ablation}. Both loops are optimized using Optuna \cite{akiba2019optuna} under a Tree-structured Parzen Estimator (TPE) sampler.

\section{Experiments and Results}
\label{sec:results}

\subsection{Experimental Setup}
\label{subsec:experimental-setup}

We evaluate FoGS on 16 public survival datasets spanning oncology, cardiology, and clinical-trial cohorts, with cohort sizes, feature counts, event rates, and missingness varying widely across the collection; aggregated per-dataset statistics are reported in Appendix~\ref{app:extended} (Table~\ref{tab:dataset-audit}). Each dataset is split into training, validation, and test with a 60-20-20 proportion, and generators and scorers are fit on the training partition only. FoGS produces $\mathcal{D}_s^{*}$ via the pipeline of Section~\ref{sec:method}, after which the inner loop trains the downstream XGBoost-Cox model on $\mathcal{D}_s^{*}$; the final model is evaluated on the held-out real test set by Harrell's C-index~\cite{harrell1996multivariable} and IBS~\cite{glenn1950verification}. Both metrics are always reported on a 0-100 scale instead of a 0-1 scale to improve readability. Synthetic-data privacy, instead, is assessed by the nearest-neighbour distance ratio (NNDR),
\begin{equation}
    \mathrm{NNDR}(x^{\mathrm{syn}}) \;=\; \frac{d_1(x^{\mathrm{syn}}, \mathcal{D}_r)}{d_2(x^{\mathrm{syn}}, \mathcal{D}_r)},
    \label{eq:nndr}
\end{equation}
the ratio of the Euclidean distances from a synthetic sample to its first- and second-nearest real neighbours in $\mathcal{D}_r$, where a lower value indicates a smaller nearest-neighbour privacy margin. The FoGS configuration adopts $\lvert\mathcal{G}\rvert = 4$ generators, $\lvert\mathcal{M}\rvert = 7$ scorers, and pool multiplier $\kappa = 100$; for each dataset we run 300 outer trials over the 21-parameter policy search space, each with 100 inner XGBoost-Cox configurations (Appendix~\ref{app:search-spaces}).

\subsection{Downstream Utility}
\label{subsec:results-utility}

\begin{figure}[t]
    \centering
    \includegraphics[width=\linewidth]{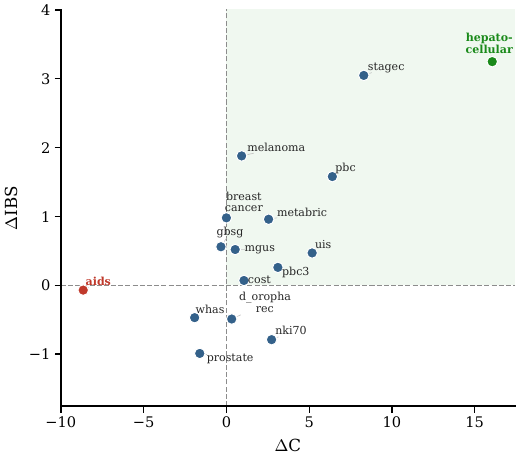}
    \caption{Per-dataset change of FoGS over the real-data baseline, $\Delta\mathrm{C}$ (horizontal) against $\Delta\mathrm{IBS}$ (vertical), one point per dataset. The shaded upper-right quadrant contains the datasets that improve on both metrics; \texttt{breast\_cancer} lies on the $\Delta\mathrm{C}=0$ axis.}
    \label{fig:scatter-dC-dIBS}
\end{figure}

Under TSTR with the XGBoost-Cox downstream model (Table~\ref{tab:results}, Fig.~\ref{fig:scatter-dC-dIBS}), FoGS improves both metrics on $9/16$ datasets and at least one on $13/16$, with a C-index tie on \texttt{breast\_cancer}; only \texttt{whas}, \texttt{aids}, and \texttt{prostate} improve on neither. Median gains are $\Delta\mathrm{C}=+0.99$ and $\Delta\mathrm{IBS}=+0.50$ (means $+2.17$ and $+0.67$). A one-sided Wilcoxon signed-rank test over the 16 paired differences gives $p=0.039$ ($\Delta\mathrm{C}$) and $p=0.035$ ($\Delta\mathrm{IBS}$).

Two ablations on the same candidate pool isolate the contribution of the generator pool and of the selection policy. Restricting FoGS to a single generator lowers aggregate utility below the full four-generator pool: the strongest single generator (ARF) reaches $\Delta\mathrm{C}=+1.48$ and $\Delta\mathrm{IBS}=+0.51$ against $+2.17$ and $+0.67$ for the pool, and the per-dataset best generator is split across all four (Tables~\ref{tab:ablation-aggregate} and~\ref{tab:ablation-cindex}). Replacing the selection policy with a random draw of equal size from the pool drops utility below the real-data baseline ($\Delta\mathrm{C}=-0.50$, $\Delta\mathrm{IBS}=-0.35$); measured directly against this random baseline, FoGS gains $\Delta\mathrm{C}=+2.67$ ($13/16$ datasets, $p=0.007$) and $\Delta\mathrm{IBS}=+1.03$ ($11/16$, $p=0.018$) (Table~\ref{tab:results_unfiltered}).

The outer objective is examined on three representative datasets (\texttt{metabric}, \texttt{pbc}, \texttt{gbsg}): minimizing $\mathrm{IBS}_{\mathrm{val}}$ yields the strongest aggregate on both metrics ($\Delta\mathrm{C}=+2.87$, $\Delta\mathrm{IBS}=+1.03$), ahead of the composite $\mathrm{C}_{\mathrm{val}}-2\,\mathrm{IBS}_{\mathrm{val}}$ ($+1.48$, $+0.78$) and of maximizing $\mathrm{C}_{\mathrm{val}}$ ($-0.55$, $-0.26$) (Table~\ref{tab:outer-ablation}).

\begin{table*}[!t]
   \centering
   \setlength{\tabcolsep}{3pt}
    \caption{Per-dataset downstream utility under TSTR: real-data baseline versus FoGS on C-index and IBS, the signed differences $\Delta$ (oriented so that a positive value denotes an improvement over the baseline for both metrics), and the FoGS NNDR. Reported values are $\times 100$. \texttt{breast\_cancer} ties on C-index and is not counted as a win. Bottom rows give the aggregates and the one-sided Wilcoxon $p$ over the 16 paired differences.}
   \label{tab:results}
   \begin{small}
   \begin{tabularx}{\textwidth}{@{}l*{7}{Y}@{}}
       \toprule
       & \multicolumn{3}{c}{C-index $\uparrow$}
       & \multicolumn{3}{c}{IBS $\downarrow$}
       & \multicolumn{1}{c}{NNDR $\uparrow$}\\
       \cmidrule(lr){2-4} \cmidrule(lr){5-7} \cmidrule(lr){8-8}
       \textbf{Dataset}
       & \textbf{Real} & \textbf{FoGS} & $\boldsymbol{\Delta}$
       & \textbf{Real} & \textbf{FoGS} & $\boldsymbol{\Delta}$
       & \textbf{FoGS} \\
       \midrule
       \texttt{gbsg}           & $\mathbf{70.05}$ & $69.72$          & \cellcolor{pred!25}$-0.33$  & $10.55$ & $\mathbf{9.99}$          & \cellcolor{pgreen!25}$+0.56$ & $74.30$ \\
       \texttt{metabric}       & $61.73$ & $\mathbf{64.28}$ & \cellcolor{pgreen!25}$+2.55$ & $10.87$ & $\mathbf{9.91}$ & \cellcolor{pgreen!25}$+0.96$ & $80.79$ \\
       \texttt{whas}           & $\mathbf{75.19}$ & $73.28$          & \cellcolor{pred!25}$-1.92$  & $\mathbf{10.61}$ & $11.08$          & \cellcolor{pred!25}$-0.47$  & $85.20$ \\
       \texttt{aids}           & $\mathbf{71.88}$ & $63.23$          & \cellcolor{pred!25}$-8.65$  & $\mathbf{4.68}$ & $4.75$          & \cellcolor{pred!25}$-0.07$  & $82.49$ \\
       \texttt{breast\_cancer} & $\mathbf{71.65}$ & $\mathbf{71.65}$          & $0.00$                      & $8.85$ & $\mathbf{7.87}$ & \cellcolor{pgreen!25}$+0.98$ & $71.41$ \\
       \texttt{cost}           & $66.39$ & $\mathbf{67.45}$ & \cellcolor{pgreen!25}$+1.06$ & $16.52$ & $\mathbf{16.45}$ & \cellcolor{pgreen!25}$+0.07$ & $57.44$ \\
       \texttt{d\_oropha\_rec} & $56.17$ & $\mathbf{56.49}$ & \cellcolor{pgreen!25}$+0.32$ & $\mathbf{15.92}$ & $16.40$          & \cellcolor{pred!25}$-0.49$  & $64.31$ \\
       \texttt{hepatocellular} & $60.86$ & $\mathbf{76.91}$ & \cellcolor{pgreen!25}$+16.06$ & $14.50$ & $\mathbf{11.25}$ & \cellcolor{pgreen!25}$+3.25$ & $84.13$ \\
       \texttt{melanoma}       & $82.02$ & $\mathbf{82.94}$ & \cellcolor{pgreen!25}$+0.92$ & $9.28$ & $\mathbf{7.41}$ & \cellcolor{pgreen!25}$+1.88$ & $71.02$ \\
       \texttt{mgus}           & $70.78$ & $\mathbf{71.31}$ & \cellcolor{pgreen!25}$+0.53$ & $11.76$ & $\mathbf{11.24}$ & \cellcolor{pgreen!25}$+0.52$ & $82.21$ \\
       \texttt{nki70}          & $71.82$ & $\mathbf{74.55}$ & \cellcolor{pgreen!25}$+2.73$ & $\mathbf{10.36}$ & $11.15$          & \cellcolor{pred!25}$-0.79$  & $70.57$ \\
       \texttt{pbc}            & $78.04$ & $\mathbf{84.43}$ & \cellcolor{pgreen!25}$+6.40$ & $8.54$ & $\mathbf{6.96}$ & \cellcolor{pgreen!25}$+1.58$ & $59.98$ \\
       \texttt{pbc3}           & $77.87$ & $\mathbf{80.98}$ & \cellcolor{pgreen!25}$+3.11$ & $6.59$ & $\mathbf{6.33}$ & \cellcolor{pgreen!25}$+0.26$ & $79.22$ \\
       \texttt{prostate}       & $\mathbf{60.45}$ & $58.84$          & \cellcolor{pred!25}$-1.61$  & $\mathbf{17.46}$ & $18.45$          & \cellcolor{pred!25}$-0.99$  & $90.50$ \\
       \texttt{stagec}         & $61.86$ & $\mathbf{70.16}$ & \cellcolor{pgreen!25}$+8.30$ & $13.32$ & $\mathbf{10.27}$ & \cellcolor{pgreen!25}$+3.05$ & $70.52$ \\
       \texttt{uis}            & $56.69$ & $\mathbf{61.87}$ & \cellcolor{pgreen!25}$+5.18$ & $10.40$ & $\mathbf{9.93}$ & \cellcolor{pgreen!25}$+0.47$ & $84.36$ \\
       \midrule
       Mean                              & -- & -- & $\boldsymbol{+2.17}$ & -- & -- & $\boldsymbol{+0.67}$ & $75.53$ \\
       Median                            & -- & -- & $\boldsymbol{+0.99}$ & -- & -- & $\boldsymbol{+0.50}$ & $76.76$ \\
       $p$ (Wilcoxon)         & -- & -- & $0.039$              & -- & -- & $0.035$              & -- \\
       Wins                              & -- & -- & $11/16$              & -- & -- & $11/16$              & -- \\
       \bottomrule
   \end{tabularx}
   \end{small}
\end{table*}

\begin{figure*}[t]
    \centering
    \includegraphics[width=\linewidth]{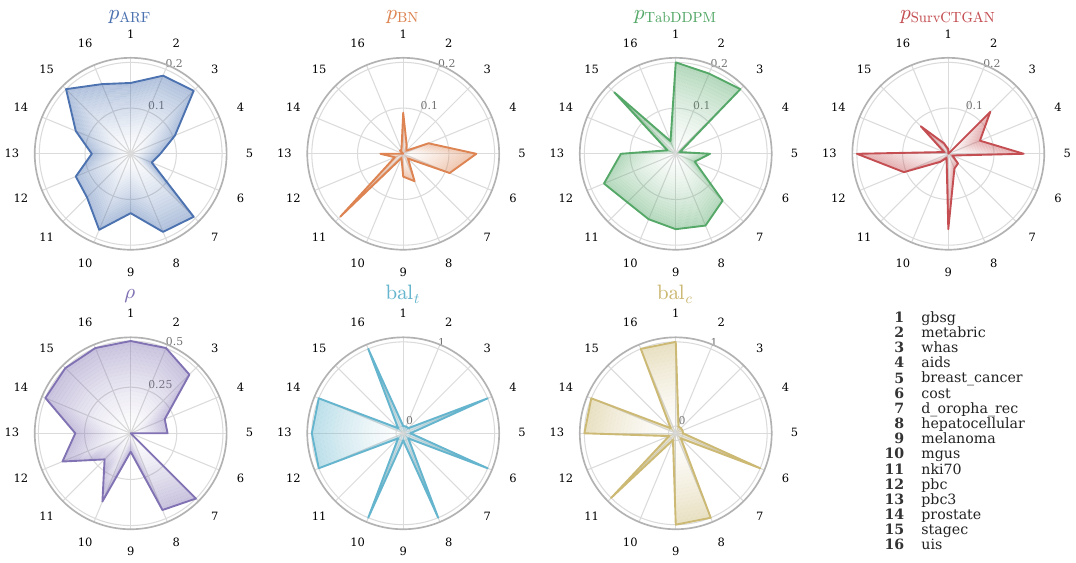}
    \caption{Best-trial selection policy for each of the 16 datasets. Top row: the four generator quotas $p_g$. Bottom row: the random complement $\rho$ and the two stratification flags $\mathrm{bal}_t$ and $\mathrm{bal}_c$. Binary panels are offset at zero for visibility.}
    \label{fig:bestpolicy-spider}
\end{figure*}

\subsection{Structure of the Selection Policy}
\label{subsec:results-policy}

The best-trial policies expose a consistent structure (Fig.~\ref{fig:bestpolicy-spider}, Table~\ref{tab:bestpolicy-full}). The random-complement fraction $\rho$ has mean $0.36$ and saturates the upper bound of its search range ($0.5$) on $6/16$ datasets, while pure top-loss selection ($\rho=0$) occurs only on \texttt{cost}; event-time and censoring stratification are each active on $8/16$ datasets.

The scorer-loss weights, by contrast, do not concentrate on any single signal (Table~\ref{tab:scorer-weights-full}, Fig.~\ref{fig:scorer-weights-heatmap}): the most frequent row-dominant combination (\texttt{DeepSurv-NLL}) leads on only $3/16$ datasets, $11$ of the $14$ combinations are row-dominant on at least one dataset, five appear among the top three weights on at least five datasets, and the cross-dataset mean weights span a narrow $0.046$--$0.089$ band.

\subsection{Nearest-Neighbour Privacy}
\label{subsec:results-privacy}

The FoGS-filtered set has mean NNDR $0.755$ (Eq.~\eqref{eq:nndr}). Relative to the unfiltered (random) draw from the generator pool, the value is higher on $12/16$ datasets (lower on the other four) and differs by $+2.82$ on average, a difference that is not significant ($p=0.087$) (Table~\ref{tab:results_unfiltered}).

\section{Discussion}
\label{sec:discussion}

Earlier survival generators produce synthetic data that underperforms real-data training under TSTR~\cite{norcliffe2023survivalgan, qian2023synthcity}. FoGS reaches or exceeds the real-data baseline on most cohorts with the same generators, improving downstream utility by selecting among their outputs rather than by changing how those outputs are produced. Two ablations on the candidate pool locate the source of this improvement. The first isolates the generator pool: the four-generator pool outperforms every single-generator restriction (Table~\ref{tab:ablation-aggregate}), and although its margin over the strongest single generator (ARF) is modest, the per-dataset best generator varies across the four (Table~\ref{tab:ablation-cindex}), so pooling removes the need to identify the most suitable generator for a cohort in advance; Appendix~\ref{app:umap} (Figs.~\ref{fig:umap-part1} and~\ref{fig:umap-part2}) shows the real, FoGS-selected, and per-generator distributions side by side for every dataset.
The second ablation isolates the selection policy: a random draw of equal size from the same pool falls below the real-data baseline ($\Delta\mathrm{C}=-0.50$), whereas the tuned policy converts the pool into a clear gain ($\Delta\mathrm{C}=+2.17$), an improvement of $+2.67$ measured directly against the random draw (Table~\ref{tab:results_unfiltered}). The improvement over real-data training therefore originates in how samples are selected rather than in the generators themselves.

The optimal policies expose a quality--coverage trade-off. On all but one dataset the best policy retains a sizeable random fraction (mean $\rho=0.36$), and selecting exclusively the most plausible samples was optimal on a single cohort only. A plausible explanation is distributional: per-sample loss is lowest in the densest, most typical regions of the covariate space, so a purely top-ranked selection over-represents those regions and under-samples the rarer cases in the tails on which a survival model relies to generalize, while the random complement restores this coverage. The mean random fraction $\rho=0.36$ stays well above zero across cohorts of widely differing size, dimensionality, and event rate suggests a broader behaviour: when synthetic samples are filtered by a plausibility score, a non-trivial degree of randomness is required to preserve population coverage.

A second pattern emerges in how the seven scorers combine in the selected policies. No fixed scorer--loss pairing prevails across datasets. In fact, we observe that no single pairing is the dominant one on more than $3/16$ cohorts, and the weights averaged over all datasets are nearly uniform (Table~\ref{tab:scorer-weights-full}). Within any single cohort, by contrast, the weight distribution is far from uniform: a few scorer--loss pairings carry most of the weight, but their identity changes from one cohort to the next. Because averaging across datasets hides these cohort-specific selections, the informative signal resides at the level of the individual dataset. This also indicates that the per-cohort tuning captures genuine structure rather than fitting validation noise. The weights consequently cannot be fixed a priori without degrading transfer to a new cohort, which is precisely why FoGS exposes them as parameters of the outer search.

The magnitude of the gain varies widely across cohorts (Fig.~\ref{fig:scatter-dC-dIBS}). Performance ranges from $+16.06$ $\Delta\mathrm{C}$ on \texttt{hepatocellular} to $-8.65$ $\Delta\mathrm{C}$ on \texttt{aids}. However, both of these extremes admit an interpretable explanation. \texttt{hepatocellular} is small ($N=227$), high-dimensional ($P=43$), and heavily incomplete ($32.07\%$ missing), a setting in which the real-data XGBoost-Cox baseline overfits and the larger, smoother synthetic pool acts as a regularizer; \texttt{aids}, by contrast, has an event rate of only $8.34\%$, so the generators observe too few events and the synthetic event-time signal degrades, biasing the downstream model toward predicting non-occurrence. To assess whether such outcomes are predictable from cohort properties, we examined the association between the per-dataset gain and cohort size, dimensionality, event rate, missingness, and real-data baseline difficulty; none showed a significant relationship with either the sign or the magnitude of the gain. The benefit of FoGS is thus broad but not deducible a priori from dataset structure.

The one-sided Wilcoxon test confirms the improvement over real-data training ($p=0.039$ for $\Delta\mathrm{C}$, $p=0.035$ for $\Delta\mathrm{IBS}$). With the two most extreme datasets (\texttt{hepatocellular} and \texttt{aids}) excluded, the C-index gain remains significant under the two-sided test ($p=0.033$). The strongest evidence, however, comes from the comparison that isolates the selection policy: against a random draw from the same pool, FoGS improves the C-index by $+2.67$ ($p=0.007$), a result that holds two-sided and remains significant under leave-one-cohort-out.

Finally, because plausibility-based selection draws the synthetic set toward the real records, and proximity to those records is what lowers the nearest-neighbour distance ratio, the fidelity gain might be expected to come at the cost of privacy margin. However, there is no significant statistical evidence for this pattern: on average the filtered set is no closer to the real cohort than an unfiltered draw ($\Delta\,\mathrm{NNDR}=+2.82$, $p=0.087$). The tuned policy assigns consistent quota to ARF (Table~\ref{tab:bestpolicy-full}), the generator that preserves the widest nearest-neighbour margin (Table~\ref{tab:ablation-aggregate}), and this allocation counterbalances the inward pull of low-loss selection. This compensation is only an aggregate effect: on the four cohorts where the filtered set sits closer to the real data than the random draw (\texttt{gbsg}, \texttt{breast\_cancer}, \texttt{cost}, \texttt{nki70}; Table~\ref{tab:results_unfiltered}), the fidelity gain does come at the cost of a narrower privacy margin.

\begin{table}[t]
\centering
\setlength{\tabcolsep}{3pt}
\caption{Aggregate utility and privacy of FoGS restricted to each single generator, of random sampling, and of the full four-generator pool. $\Delta\mathrm{C}$ and $\Delta\mathrm{IBS}$ are the mean per-dataset gains over the real-data baseline ($\times 100$); NNDR is the cohort-mean nearest-neighbour distance ratio.}
\label{tab:ablation-aggregate}
\begin{small}
\begin{tabularx}{\columnwidth}{@{}lYYY@{}}
\toprule
\textbf{Variant} 
& $\boldsymbol{\Delta\mathrm{C}} \uparrow$ 
& $\boldsymbol{\Delta\mathrm{IBS}} \uparrow$ 
& \textbf{NNDR} $\uparrow$ \\
\midrule
ARF 
& $+1.48$ 
& $+0.51$ 
& $\mathbf{89.2}$ \\

BN 
& $-1.64$ 
& $+0.14$ 
& $43.7$ \\

TabDDPM 
& $-0.44$ 
& $0.00$ 
& $73.0$ \\

SurvCTGAN 
& $-3.95$ 
& $-1.20$ 
& $85.4$ \\

\midrule
Random Sampling 
& $-0.50$ 
& $-0.35$ 
& $72.7$ \\

\midrule
\textbf{FoGS} 
& $\boldsymbol{+2.17}$ 
& $\boldsymbol{+0.67}$ 
& $75.5$ \\
\bottomrule
\end{tabularx}
\end{small}
\end{table}

\section{Limitations and Future Directions}
\label{sec:limitations}

\textbf{Generator quality on hard cohorts.} Three of the 16 datasets (\texttt{whas}, \texttt{aids}, \texttt{prostate}) improve on neither metric, for two distinct reasons. The first is the capacity of the generators: in aggregate, only ARF yields an improvement over the real-data baseline among the single generators (Table~\ref{tab:ablation-aggregate}), and on \texttt{whas} and \texttt{prostate} even the full pool does not reproduce the data distribution closely enough to overtake real-data training. The second reason is intrinsic to the cohort: \texttt{aids} couples a very low event rate ($8.34\%$) with correspondingly heavy censoring, so the binding factor is the scarcity of observed events in the data itself rather than the fidelity of any generator.

\textbf{A single optimization run per dataset.} Each result is obtained from one outer-Optuna trajectory. The full pipeline couples four-generator and seven-scorer pretraining with a 300-trial policy search, which makes repeated runs across random seeds computationally impractical; we therefore report a single seeded run per dataset rather than a variance estimate over seeds. The fixed seed makes each reported result exactly reproducible, and quantifying the sensitivity of the per-dataset estimates to the seed is left to future work.

\textbf{The policy search bounds the random fraction.} The random complement $\rho$ is searched only up to $0.5$, and on 6 of the 16 datasets the optimum saturates this bound. On those cohorts the true optimum may therefore lie beyond $0.5$, and because their values are truncated at the bound, the cross-dataset mean of $0.36$ understates the randomness the policy favours and should be read as a lower bound.

\textbf{A single downstream model.} The selection policy is tuned exclusively against XGBoost-Cox. Although XGBoost-Cox is a strong, state-of-the-art tabular survival model, it remains a single choice of downstream learner: early experiments on a subset of datasets showed no significant difference in downstream model selection, but the tuned policy may still carry a bias toward XGBoost-Cox in the full experiment set.

\textbf{Computational cost of the nested search.} Because the two levels are nested, the cost of a single run is the product of the two trial budgets: every outer trial launches a full inner hyperparameter search to obtain an unbiased validation signal for its candidate policy, so the $300$ outer and $100$ inner trials amount to roughly $3\times10^{4}$ downstream XGBoost-Cox fits per dataset. This is incurred on top of the one-time pretraining of the four generators and seven scorers and the scoring of the $\kappa\lvert\mathcal{G}\rvert N_r$ candidate pool. The inner validation loop is the bottleneck and cannot be bypassed without biasing the outer objective. Cheaper surrogates for the inner loop would lower the cost, but each weakens the validation signal that drives the selection-policy search.

\textbf{A single privacy proxy.} Privacy is assessed only through the nearest-neighbour distance ratio. Membership-inference and attribute-inference attacks, and formal differential-privacy guarantees, lie outside the scope of this work.

Several of these limitations point directly to future work. The pool accepts new generators without any change to the pipeline, so the framework can track improvements in tabular generation and incorporate foundation-model-based survival scorers as they mature. Widening the policy search, and the range of $\rho$ in particular, would locate the true coverage optimum on the cohorts that currently reach the bound. Because the framework does not depend on the downstream task, it extends beyond survival to other tabular clinical problems, including regression, class-imbalanced classification, and competing-risks modelling, as well as to other downstream survival models. Finally, a fuller privacy assessment, with inference attacks and formal guarantees, would define the conditions under which the released cohorts can be shared safely.

\section{Conclusions}
\label{sec:conclusion}

We presented FoGS, a pipeline that builds a synthetic survival training set by pooling four tabular generators, scoring every candidate sample with seven survival models trained on real data, and tuning the selection policy---generator quotas, scorer weights, stratification, and a random complement---against held-out real-validation performance through a two-level optimization. Across 16 public datasets FoGS matches or exceeds real-data training on most cohorts, and its comparison against random selection shows that the improvement comes from how the samples are selected rather than from the generators themselves. Two findings recur across datasets: a quality--coverage trade-off, which requires retaining a non-trivial random fraction so that the rare cases in the tails are not discarded, and a dataset-conditional structure in the scorer combination, under which the weighting of plausibility signals must be tuned per cohort rather than fixed in advance. The filtered cohorts show no significant change in nearest-neighbour privacy margin relative to unfiltered sampling. Because it requires only publicly available generators and standard survival scorers, FoGS can be directly applied to new cohorts.

% \section*{Acknowledgment}
% Placeholder: funding and institutional affiliations to be completed by the authors.

\bibliographystyle{plainnat}
\bibliography{biblio}

\clearpage
\appendix

\section{Search Spaces and Per-Sample Loss Formulas}
\label{app:search-spaces}

\subsection{Outer-Optuna Search Space}
The 21 policy parameters are jointly optimized over 300 trials per dataset by a TPE multivariate sampler with constant-liar for parallel-worker safety, seeded for reproducibility (Table~\ref{tab:hyperparams-pipeline}).

\subsection{Inner-Optuna Search Space}
The downstream XGBoost-Cox model (objective \texttt{survival:\allowbreak cox}) is tuned over 100 trials per outer trial against $\mathrm{C} - 2\,\mathrm{IBS}$ on real validation, with a fixed per-dataset seed (Table~\ref{tab:inner-hps}).

\subsection{Per-Sample Loss Formulas}
For a synthetic sample $i = (\mathbf{x}_i, t_i, \delta_i)$ scored by model $m$ with predicted survival function $S_m(\cdot \mid \mathbf{x}_i)$, evaluated on a time grid $\{\tau_1, \dots, \tau_K\}$ spanning $[0, \max(t_{\mathrm{real}})]$, the per-sample integrated Brier score is
\begin{align}
    \mathrm{IBS}_{m,i} &= \frac{1}{K}\sum_{j=1}^{K} \mathrm{BS}_{m,i}(\tau_j), \label{eq:ibs_per_sample} \\
    \mathrm{BS}_{m,i}(\tau_j) &= \mathbb{1}(t_i \le \tau_j, \delta_i = 1)\, S_m(\tau_j \mid \mathbf{x}_i)^2 \notag \\
    &\quad + \mathbb{1}(t_i > \tau_j)\,\big(1 - S_m(\tau_j \mid \mathbf{x}_i)\big)^2. \notag
\end{align}
Inverse probability of censoring weighting (IPCW) is omitted to prevent inconsistent reweighting from repeated synthetic data sampling. The per-sample negative log-likelihood is
\begin{equation}
    \mathrm{NLL}_{m,i} = -\big[\, \delta_i \log h_m(t_i \mid \mathbf{x}_i) - H_m(t_i \mid \mathbf{x}_i) \,\big],
    \label{eq:nll_per_sample}
\end{equation}
where $h_m$ and $H_m$ are the hazard and cumulative hazard predicted by model $m$; for models providing only $S_m$, these are computed by finite-difference approximation on the time grid.

\begin{table}[t]
\centering
\setlength{\tabcolsep}{3pt}
\caption{Outer-Optuna search space (21 parameters). The 14 scorer weights are normalized by their sum, as in Eq.~\eqref{eq:perSampleLoss}.}
\label{tab:hyperparams-pipeline}
\small
\begin{tabularx}{\columnwidth}{@{}lllX@{}}
\toprule
\textbf{Parameter} 
& \textbf{Domain} 
& \textbf{Range} 
& \textbf{Interpretation} \\
\midrule
$p_{\mathrm{ARF}}$ 
& continuous 
& $[0, 0.20]$ 
& ARF quota \\

$p_{\mathrm{BN}}$ 
& continuous 
& $[0, 0.20]$ 
& Bayesian Network quota \\

$p_{\mathrm{TabDDPM}}$ 
& continuous 
& $[0, 0.20]$ 
& TabDDPM quota \\

$p_{\mathrm{SurvCTGAN}}$ 
& continuous 
& $[0, 0.20]$ 
& SurvivalCTGAN quota \\

$w_{m,\mathrm{IBS}}$ 
& continuous 
& $[0, 1]$ 
& 7 IBS weights \\

$w_{m,\mathrm{NLL}}$ 
& continuous 
& $[0, 1]$ 
& 7 NLL weights \\

$\rho$ 
& continuous 
& $[0, 0.5]$ 
& random complement \\

$\mathrm{bal}_t$ 
& binary 
& $\{0, 1\}$ 
& event-time stratification \\

$\mathrm{bal}_c$ 
& binary 
& $\{0, 1\}$ 
& censoring stratification \\
\bottomrule
\end{tabularx}
\end{table}

\begin{table}[t]
\centering
\setlength{\tabcolsep}{3pt}
\caption{Inner-Optuna search space for the downstream XGBoost-Cox model, identical for FoGS and the real-data baseline.}
\label{tab:inner-hps}
\small
\begin{tabularx}{\columnwidth}{@{}lllX@{}}
\toprule
\textbf{Parameter} 
& \textbf{Domain} 
& \textbf{Range} 
& \textbf{Sampling} \\
\midrule
\texttt{learning\_rate} 
& continuous 
& $[0.01, 0.3]$ 
& log-uniform \\

\texttt{max\_depth} 
& integer 
& $[3, 10]$ 
& uniform \\

\texttt{n\_estimators} 
& integer 
& $[50, 500]$ 
& uniform (step 25) \\

\texttt{subsample} 
& continuous 
& $[0.5, 1.0]$ 
& uniform \\

\texttt{colsample\_bytree} 
& continuous 
& $[0.5, 1.0]$ 
& uniform \\

\texttt{gamma} 
& continuous 
& $[0, 5]$ 
& uniform \\

\texttt{min\_child\_weight} 
& integer 
& $[1, 10]$ 
& uniform \\

\texttt{reg\_alpha} 
& continuous 
& $[10^{-3}, 10]$ 
& log-uniform \\

\texttt{reg\_lambda} 
& continuous 
& $[10^{-3}, 10]$ 
& log-uniform \\
\bottomrule
\end{tabularx}
\end{table}

\section{Outer-Objective Ablation}
\label{app:outer-ablation}

Three representative datasets (\texttt{metabric}, \texttt{pbc}, \texttt{gbsg}) are evaluated under three outer-Optuna objectives: minimize $\mathrm{IBS}_{\mathrm{val}}$ (the main pipeline), maximize the composite $\mathrm{C}_{\mathrm{val}} - 2\,\mathrm{IBS}_{\mathrm{val}}$, and maximize $\mathrm{C}_{\mathrm{val}}$. Each cell uses 300 outer trials with inner-Optuna XGBoost-Cox tuning identical to the search space of Appendix~\ref{app:search-spaces} (Table~\ref{tab:outer-ablation}).

\begin{table*}[t]
\centering
\setlength{\tabcolsep}{4pt}
\caption{Outer-objective ablation on three datasets under three outer objectives ($300$ trials each): per-dataset real and FoGS values with the signed differences $\Delta$ ($\times 100$), and the per-objective means.}
\label{tab:outer-ablation}
\small
\begin{tabularx}{\textwidth}{@{}>{\raggedright\arraybackslash}l l Y Y Y Y Y Y@{}}
\toprule
\textbf{Outer objective} 
& \textbf{Dataset} 
& \textbf{C-Real} 
& \textbf{C-FoGS} 
& $\boldsymbol{\Delta\mathrm{C}}$ 
& \textbf{IBS-Real} 
& \textbf{IBS-FoGS} 
& $\boldsymbol{\Delta\mathrm{IBS}}$ \\
\midrule

Min $\mathrm{IBS}_{\mathrm{val}}$ (main) 
& \texttt{gbsg}     
& $70.05$ 
& $69.72$ 
& $-0.33$ 
& $10.55$ 
& $9.99$ 
& $+0.56$ \\

Min $\mathrm{IBS}_{\mathrm{val}}$ (main) 
& \texttt{metabric} 
& $61.73$ 
& $64.28$ 
& $+2.55$ 
& $10.87$ 
& $9.91$ 
& $+0.96$ \\

Min $\mathrm{IBS}_{\mathrm{val}}$ (main) 
& \texttt{pbc}      
& $78.04$ 
& $84.43$ 
& $+6.40$ 
& $8.54$ 
& $6.96$ 
& $+1.58$ \\

\textit{Mean} (Min $\mathrm{IBS}_{\mathrm{val}}$) 
& - 
& - 
& - 
& $\boldsymbol{+2.87}$ 
& - 
& - 
& $\boldsymbol{+1.03}$ \\

\midrule

Max $\mathrm{C}_{\mathrm{val}} - 2\,\mathrm{IBS}_{\mathrm{val}}$ 
& \texttt{gbsg}     
& $70.05$ 
& $70.87$ 
& $+0.82$ 
& $10.55$ 
& $9.58$ 
& $+0.97$ \\

Max $\mathrm{C}_{\mathrm{val}} - 2\,\mathrm{IBS}_{\mathrm{val}}$ 
& \texttt{metabric} 
& $61.73$ 
& $61.36$ 
& $-0.37$ 
& $10.87$ 
& $10.67$ 
& $+0.20$ \\

Max $\mathrm{C}_{\mathrm{val}} - 2\,\mathrm{IBS}_{\mathrm{val}}$ 
& \texttt{pbc}      
& $78.04$ 
& $82.04$ 
& $+4.00$ 
& $8.54$ 
& $7.37$ 
& $+1.17$ \\

\textit{Mean} (composite) 
& - 
& - 
& - 
& $+1.48$ 
& - 
& - 
& $+0.78$ \\

\midrule

Max $\mathrm{C}_{\mathrm{val}}$ 
& \texttt{gbsg}     
& $70.05$ 
& $68.06$ 
& $-1.98$ 
& $10.55$ 
& $11.02$ 
& $-0.48$ \\

Max $\mathrm{C}_{\mathrm{val}}$ 
& \texttt{metabric} 
& $61.73$ 
& $62.80$ 
& $+1.08$ 
& $10.87$ 
& $10.75$ 
& $+0.12$ \\

Max $\mathrm{C}_{\mathrm{val}}$ 
& \texttt{pbc}      
& $78.04$ 
& $77.29$ 
& $-0.75$ 
& $8.54$ 
& $8.96$ 
& $-0.42$ \\

\textit{Mean} (Max $\mathrm{C}_{\mathrm{val}}$) 
& - 
& - 
& - 
& $-0.55$ 
& - 
& - 
& $-0.26$ \\

\bottomrule
\end{tabularx}
\end{table*}

Minimizing $\mathrm{IBS}_{\mathrm{val}}$ aggregates to $+2.87$ $\Delta\mathrm{C}$ and $+1.03$ $\Delta\mathrm{IBS}$, dominating both alternatives on both metrics. The composite objective produces a moderate $\Delta\mathrm{C}$ gain but underperforms $\mathrm{IBS}_{\mathrm{val}}$ on calibration. Maximizing $\mathrm{C}_{\mathrm{val}}$ is the worst configuration, producing a net negative aggregate on both metrics, consistent with rank-based optimization being sensitive to validation-set finite-sample noise.

\section{Datasets and Extended Per-Dataset Results}
\label{app:extended}
\begin{figure}[t]
    \centering
    \includegraphics[width=\linewidth]{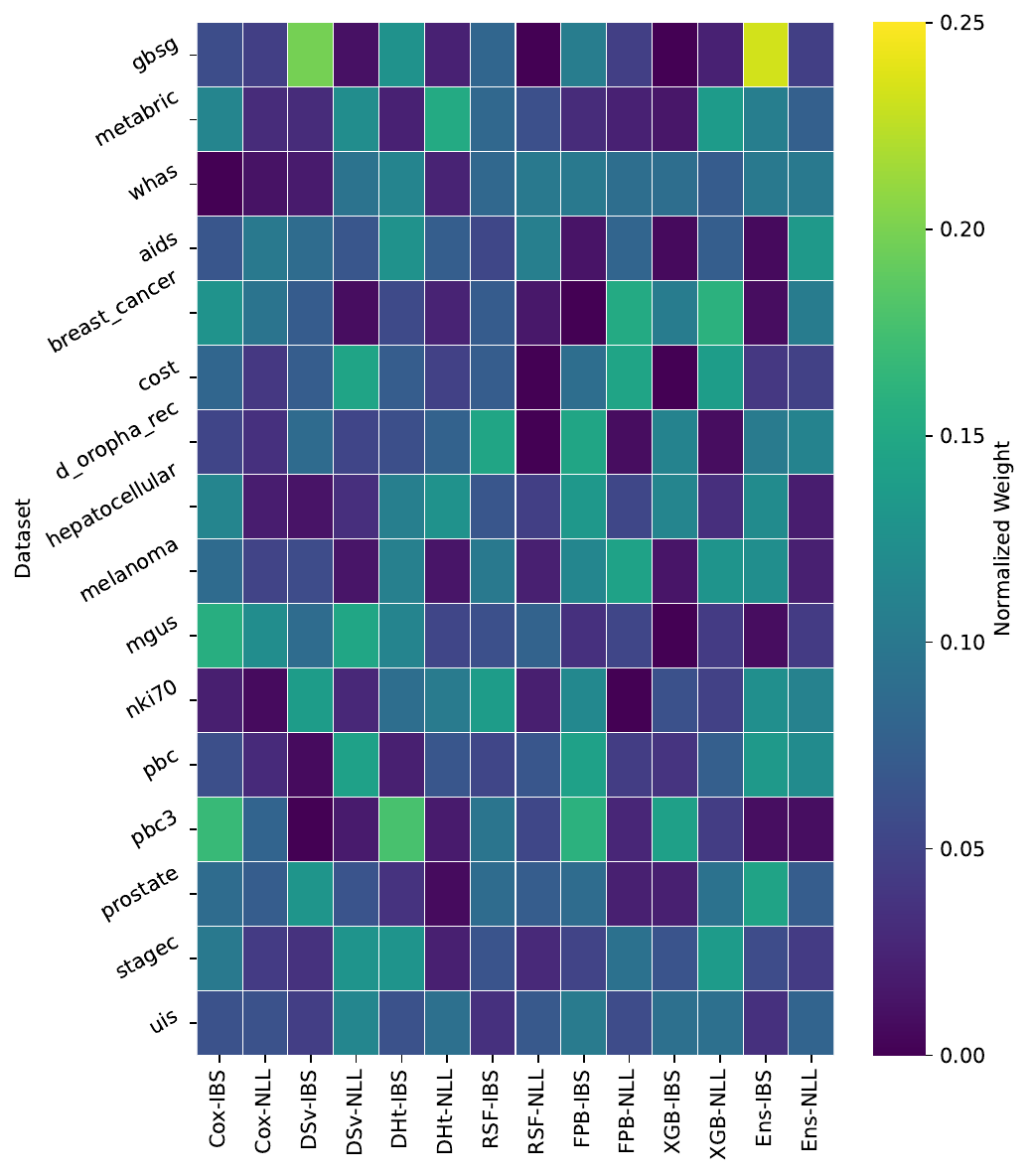}
    \caption{Best-trial scorer-loss weights for each dataset over the 14 combinations (columns). Colour encodes the weight; each row sums to one.}
    \label{fig:scorer-weights-heatmap}
\end{figure}

Table~\ref{tab:dataset-audit} reports aggregated statistics for the 16 datasets.
Table~\ref{tab:bestpolicy-full} is the numeric expansion of Fig.~\ref{fig:bestpolicy-spider}.
Table~\ref{tab:scorer-weights-full} and Fig.~\ref{fig:scorer-weights-heatmap} report the normalized best-trial scorer-loss weights; per-dataset NNDR for the filtered and random sets is included in Table~\ref{tab:results_unfiltered}.

\begin{table}[t]
\centering
\caption{Descriptors for the 16 survival datasets: cohort size $N$, feature count $P$, event and censoring rates (\%) and missingness (\%).}
\label{tab:dataset-audit}
\footnotesize
\begin{tabular}{@{}lrrccc@{}}
\toprule
\textbf{Dataset} & \textbf{N} & \textbf{P} & \textbf{event \%} & \textbf{cens. \%} & \textbf{miss. \%} \\
\midrule
\texttt{gbsg}           & 686 & 8 & 43.59 & 56.41 & 0.00 \\
\texttt{metabric}       & 1904 & 9 & 57.93 & 42.07 & 0.00 \\
\texttt{whas}           & 461 & 16 & 38.18 & 61.82 & 0.00 \\
\texttt{aids}           & 1151 & 11 & 8.34 & 91.66 & 0.00 \\
\texttt{breast\_cancer} & 198 & 80 & 25.76 & 74.24 & 0.00 \\
\texttt{cost}           & 518 & 13 & 77.99 & 22.01 & 0.00 \\
\texttt{d\_oropha\_rec} & 192 & 12 & 72.40 & 27.60 & 0.00 \\
\texttt{hepatocellular} & 227 & 43 & 42.73 & 57.27 & 32.07 \\
\texttt{melanoma}       & 205 & 5 & 27.80 & 72.20 & 0.00 \\
\texttt{mgus}           & 241 & 9 & 93.36 & 6.64 & 9.51 \\
\texttt{nki70}          & 144 & 75 & 33.33 & 66.67 & 0.00 \\
\texttt{pbc}            & 312 & 6 & 40.06 & 59.94 & 0.00 \\
\texttt{pbc3}           & 349 & 12 & 17.48 & 82.52 & 0.53 \\
\texttt{prostate}       & 502 & 15 & 70.52 & 29.48 & 0.32 \\
\texttt{stagec}         & 146 & 7 & 36.99 & 63.01 & 0.76 \\
\texttt{uis}            & 628 & 8 & 80.89 & 19.11 & 0.88 \\
\bottomrule
\end{tabular}
\end{table}

\begin{table*}[t]
\centering
\caption{Best-trial selection-policy parameters per dataset: generator quotas $p_g$ (fractions of $N_g$), the random complement $\rho$, and the stratification flags $\mathrm{bal}_t$, $\mathrm{bal}_c$ (\cmark{} active, \xmark{} inactive).}
\label{tab:bestpolicy-full}
\begin{tabularx}{\linewidth}{@{}lYYYYYYY@{}}
\toprule
\textbf{Dataset} & $p_{\mathrm{ARF}}$ & $p_{\mathrm{BN}}$ & $p_{\mathrm{TabDDPM}}$ & $p_{\mathrm{SurvCTGAN}}$ & $\rho$ & $\mathrm{bal}_t$ & $\mathrm{bal}_c$ \\
\midrule
\texttt{gbsg}           & $0.155$ & $0.090$ & $0.200$ & $0.010$ & $0.50$ & \xmark & \cmark \\
\texttt{metabric}       & $0.185$ & $0.020$ & $0.190$ & $0.000$ & $0.50$ & \xmark & \xmark \\
\texttt{whas}           & $0.195$ & $0.010$ & $0.200$ & $0.130$ & $0.45$ & \xmark & \xmark \\
\texttt{aids}           & $0.105$ & $0.060$ & $0.010$ & $0.075$ & $0.20$ & \cmark & \xmark \\
\texttt{breast\_cancer} & $0.065$ & $0.160$ & $0.075$ & $0.165$ & $0.20$ & \xmark & \xmark \\
\texttt{cost}           & $0.050$ & $0.110$ & $0.045$ & $0.010$ & $0.00$ & \cmark & \cmark \\
\texttt{d\_oropha\_rec} & $0.195$ & $0.015$ & $0.145$ & $0.030$ & $0.50$ & \xmark & \xmark \\
\texttt{hepatocellular} & $0.185$ & $0.065$ & $0.170$ & $0.035$ & $0.45$ & \cmark & \cmark \\
\texttt{melanoma}       & $0.130$ & $0.050$ & $0.165$ & $0.165$ & $0.10$ & \xmark & \cmark \\
\texttt{mgus}           & $0.180$ & $0.010$ & $0.155$ & $0.010$ & $0.40$ & \cmark & \xmark \\
\texttt{nki70}          & $0.135$ & $0.195$ & $0.150$ & $0.025$ & $0.20$ & \xmark & \cmark \\
\texttt{pbc}            & $0.130$ & $0.020$ & $0.170$ & $0.105$ & $0.40$ & \cmark & \xmark \\
\texttt{pbc3}           & $0.085$ & $0.050$ & $0.120$ & $0.200$ & $0.30$ & \cmark & \cmark \\
\texttt{prostate}       & $0.130$ & $0.000$ & $0.020$ & $0.010$ & $0.50$ & \cmark & \cmark \\
\texttt{stagec}         & $0.200$ & $0.010$ & $0.190$ & $0.085$ & $0.50$ & \xmark & \xmark \\
\texttt{uis}            & $0.165$ & $0.010$ & $0.030$ & $0.025$ & $0.50$ & \cmark & \cmark \\
\midrule
\textbf{Mean / Count}   & $\mathbf{0.143}$ & $\mathbf{0.055}$ & $\mathbf{0.127}$ & $\mathbf{0.068}$ & $\mathbf{0.356}$ & $\mathbf{8/16}$ & $\mathbf{8/16}$ \\
\bottomrule
\end{tabularx}
\end{table*}

\begin{table*}[t]
\centering
\caption{Best-trial scorer-loss weights per dataset, normalized to row-sum one; bold marks the largest weight in each row. Each scorer (Cox, DSv = DeepSurv, DHt = DeepHit, RSF, FPb = FPBoost, XGB = XGBoost-Cox, Ens = Ensemble) is paired with IBS and NLL.}
\label{tab:scorer-weights-full}
\resizebox{\textwidth}{!}{%
\begin{tabular}{@{}lcccccccccccccc@{}}
\toprule
\textbf{Dataset} & \textbf{Cox-IBS} & \textbf{Cox-NLL} & \textbf{DSv-IBS} & \textbf{DSv-NLL} & \textbf{DHt-IBS} & \textbf{DHt-NLL} & \textbf{RSF-IBS} & \textbf{RSF-NLL} & \textbf{FPb-IBS} & \textbf{FPb-NLL} & \textbf{XGB-IBS} & \textbf{XGB-NLL} & \textbf{Ens-IBS} & \textbf{Ens-NLL} \\
\midrule
\texttt{gbsg}           & $0.058$ & $0.047$ & $0.198$ & $0.012$ & $0.128$ & $0.023$ & $0.081$ & $0.000$ & $0.105$ & $0.047$ & $0.000$ & $0.023$ & $\mathbf{0.233}$ & $0.047$ \\
\texttt{metabric}       & $0.114$ & $0.030$ & $0.030$ & $0.121$ & $0.023$ & $\mathbf{0.152}$ & $0.083$ & $0.061$ & $0.030$ & $0.023$ & $0.015$ & $0.136$ & $0.106$ & $0.076$ \\
\texttt{whas}           & $0.000$ & $0.012$ & $0.018$ & $0.095$ & $\mathbf{0.113}$ & $0.024$ & $0.083$ & $0.101$ & $0.101$ & $0.089$ & $0.089$ & $0.071$ & $0.101$ & $0.101$ \\
\texttt{aids}           & $0.067$ & $0.101$ & $0.087$ & $0.067$ & $0.128$ & $0.074$ & $0.054$ & $0.107$ & $0.013$ & $0.081$ & $0.007$ & $0.074$ & $0.007$ & $\mathbf{0.134}$ \\
\texttt{breast\_cancer} & $0.128$ & $0.096$ & $0.072$ & $0.008$ & $0.056$ & $0.024$ & $0.072$ & $0.016$ & $0.000$ & $0.152$ & $0.104$ & $\mathbf{0.160}$ & $0.008$ & $0.104$ \\
\texttt{cost}           & $0.081$ & $0.041$ & $0.073$ & $\mathbf{0.146}$ & $0.073$ & $0.049$ & $0.073$ & $0.000$ & $0.089$ & $0.146$ & $0.000$ & $0.138$ & $0.041$ & $0.049$ \\
\texttt{d\_oropha\_rec} & $0.052$ & $0.034$ & $0.086$ & $0.052$ & $0.060$ & $0.078$ & $\mathbf{0.147}$ & $0.000$ & $0.147$ & $0.009$ & $0.112$ & $0.009$ & $0.103$ & $0.112$ \\
\texttt{hepatocellular} & $0.113$ & $0.020$ & $0.013$ & $0.033$ & $0.107$ & $0.127$ & $0.067$ & $0.047$ & $\mathbf{0.133}$ & $0.053$ & $0.113$ & $0.033$ & $0.120$ & $0.020$ \\
\texttt{melanoma}       & $0.086$ & $0.050$ & $0.058$ & $0.014$ & $0.108$ & $0.014$ & $0.101$ & $0.022$ & $0.115$ & $\mathbf{0.144}$ & $0.014$ & $0.129$ & $0.122$ & $0.022$ \\
\texttt{mgus}           & $\mathbf{0.157}$ & $0.122$ & $0.087$ & $0.148$ & $0.113$ & $0.052$ & $0.061$ & $0.078$ & $0.035$ & $0.052$ & $0.000$ & $0.043$ & $0.009$ & $0.043$ \\
\texttt{nki70}          & $0.021$ & $0.007$ & $\mathbf{0.137}$ & $0.027$ & $0.089$ & $0.103$ & $\mathbf{0.137}$ & $0.021$ & $0.116$ & $0.000$ & $0.062$ & $0.048$ & $0.123$ & $0.110$ \\
\texttt{pbc}            & $0.060$ & $0.030$ & $0.007$ & $\mathbf{0.142}$ & $0.022$ & $0.067$ & $0.052$ & $0.067$ & $\mathbf{0.142}$ & $0.045$ & $0.037$ & $0.075$ & $0.134$ & $0.119$ \\
\texttt{pbc3}           & $0.168$ & $0.080$ & $0.000$ & $0.018$ & $\mathbf{0.177}$ & $0.018$ & $0.097$ & $0.053$ & $0.159$ & $0.027$ & $0.142$ & $0.044$ & $0.009$ & $0.009$ \\
\texttt{prostate}       & $0.087$ & $0.072$ & $0.130$ & $0.065$ & $0.036$ & $0.007$ & $0.087$ & $0.072$ & $0.087$ & $0.022$ & $0.022$ & $0.094$ & $\mathbf{0.145}$ & $0.072$ \\
\texttt{stagec}         & $0.101$ & $0.043$ & $0.036$ & $0.129$ & $0.129$ & $0.022$ & $0.065$ & $0.029$ & $0.050$ & $0.094$ & $0.065$ & $\mathbf{0.137}$ & $0.058$ & $0.043$ \\
\texttt{uis}            & $0.063$ & $0.063$ & $0.046$ & $\mathbf{0.114}$ & $0.063$ & $0.091$ & $0.034$ & $0.069$ & $0.103$ & $0.057$ & $0.091$ & $0.091$ & $0.034$ & $0.080$ \\
\midrule
\textbf{Mean}           & $0.085$ & $0.053$ & $0.067$ & $0.075$ & $0.089$ & $0.058$ & $0.081$ & $0.046$ & $0.089$ & $0.065$ & $0.055$ & $0.082$ & $0.085$ & $0.071$ \\
\bottomrule
\end{tabular}%
}
\end{table*}

\begin{table}[t]
\centering
\caption{Per-dataset C-index $\times 100$ of FoGS restricted to each single generator; bold identifies the best model per dataset.}
\label{tab:ablation-cindex}
\begin{tabular}{@{}lcccc@{}}
\toprule
\textbf{Dataset} & ARF & BN & TabDDPM & SurvCTGAN \\
\midrule
\texttt{gbsg} & $65.70$ & $64.69$ & $\mathbf{70.30}$ & $67.99$ \\
\texttt{metabric} & $61.10$ & $58.18$ & $60.63$ & $\mathbf{64.19}$ \\
\texttt{whas} & $73.78$ & $67.08$ & $\mathbf{76.88}$ & $71.96$ \\
\texttt{aids} & $\mathbf{70.89}$ & $59.99$ & $69.01$ & $63.87$ \\
\texttt{breast\_cancer} & $67.05$ & $\mathbf{69.35}$ & $64.94$ & $29.12$ \\
\texttt{cost} & $\mathbf{66.31}$ & $65.22$ & $64.95$ & $59.25$ \\
\texttt{d\_oropha\_rec} & $56.82$ & $\mathbf{65.26}$ & $62.18$ & $54.46$ \\
\texttt{hepatocellular} & $\mathbf{77.22}$ & $61.54$ & $69.72$ & $72.02$ \\
\texttt{melanoma} & $\mathbf{83.73}$ & $82.02$ & $76.25$ & $76.12$ \\
\texttt{mgus} & $68.65$ & $\mathbf{70.87}$ & $70.74$ & $70.56$ \\
\texttt{nki70} & $73.18$ & $\mathbf{83.64}$ & $66.14$ & $67.73$ \\
\texttt{pbc} & $\mathbf{82.46}$ & $78.46$ & $79.16$ & $78.68$ \\
\texttt{pbc3} & $\mathbf{78.85}$ & $72.30$ & $72.79$ & $\mathbf{78.85}$ \\
\texttt{prostate} & $\mathbf{59.89}$ & $56.93$ & $55.54$ & $56.81$ \\
\texttt{stagec} & $\mathbf{69.17}$ & $51.78$ & $66.80$ & $60.47$ \\
\texttt{uis} & $\mathbf{62.33}$ & $59.83$ & $60.32$ & $58.15$ \\
\bottomrule
\end{tabular}
\end{table}

\begin{table}[!t]
\centering
\setlength{\tabcolsep}{3pt}
\caption{Per-dataset IBS $\times 100$ of FoGS restricted to each single generator; bold identifies the best (lowest IBS) model per dataset.}
\label{tab:ablation-ibs}
\small
\begin{tabularx}{\columnwidth}{@{}lYYcc@{}}
\toprule
\textbf{Dataset} & \textbf{ARF} & \textbf{BN} & \textbf{TabDDPM} & \textbf{SurvCTGAN} \\
\midrule
\texttt{gbsg} & $10.69$ & $\mathbf{10.49}$ & $10.65$ & $11.85$ \\
\texttt{metabric} & $10.42$ & $11.24$ & $10.79$ & $\mathbf{10.40}$ \\
\texttt{whas} & $10.23$ & $13.28$ & $\mathbf{10.20}$ & $11.49$ \\
\texttt{aids} & $\mathbf{4.27}$ & $4.81$ & $4.86$ & $5.21$ \\
\texttt{breast\_cancer} & $8.04$ & $\mathbf{7.92}$ & $8.84$ & $14.17$ \\
\texttt{cost} & $\mathbf{16.78}$ & $16.86$ & $16.86$ & $19.70$ \\
\texttt{d\_oropha\_rec} & $14.52$ & $14.76$ & $\mathbf{13.24}$ & $15.03$ \\
\texttt{hepatocellular} & $14.01$ & $13.73$ & $18.03$ & $\mathbf{13.34}$ \\
\texttt{melanoma} & $9.12$ & $\mathbf{8.73}$ & $8.81$ & $8.74$ \\
\texttt{mgus} & $11.21$ & $11.21$ & $\mathbf{10.91}$ & $11.50$ \\
\texttt{nki70} & $11.17$ & $\mathbf{7.75}$ & $10.82$ & $10.70$ \\
\texttt{pbc} & $\mathbf{7.69}$ & $9.00$ & $7.95$ & $8.85$ \\
\texttt{pbc3} & $6.70$ & $7.78$ & $7.21$ & $\mathbf{6.61}$ \\
\texttt{prostate} & $\mathbf{16.57}$ & $17.22$ & $18.54$ & $18.37$ \\
\texttt{stagec} & $\mathbf{10.82}$ & $13.13$ & $12.50$ & $22.27$ \\
\texttt{uis} & $\mathbf{9.75}$ & $10.03$ & $10.04$ & $11.13$ \\
\bottomrule
\end{tabularx}
\end{table}

\begin{table}[!t]
\centering
\setlength{\tabcolsep}{3pt}
\caption{Per-dataset NNDR of FoGS restricted to each single generator; bold marks the highest value per dataset.}
\label{tab:ablation-nndr}
\small
\begin{tabularx}{\columnwidth}{@{}lYYcc@{}}
\toprule
\textbf{Dataset} & \textbf{ARF} & \textbf{BN} & \textbf{TabDDPM} & \textbf{SurvCTGAN} \\
\midrule
\texttt{gbsg} & $\mathbf{87.30}$ & $51.70$ & $77.70$ & $83.30$ \\
\texttt{metabric} & $\mathbf{86.70}$ & $46.70$ & $80.30$ & $85.20$ \\
\texttt{whas} & $\mathbf{91.50}$ & $41.00$ & $77.10$ & $89.80$ \\
\texttt{aids} & $\mathbf{89.10}$ & $75.50$ & $72.10$ & $84.50$ \\
\texttt{breast\_cancer} & $96.70$ & $23.80$ & $\mathbf{98.70}$ & $97.20$ \\
\texttt{cost} & $\mathbf{90.00}$ & $39.10$ & $65.10$ & $84.00$ \\
\texttt{d\_oropha\_rec} & $\mathbf{91.30}$ & $22.20$ & $27.70$ & $84.40$ \\
\texttt{hepatocellular} & $95.10$ & $20.60$ & $\mathbf{97.20}$ & $91.40$ \\
\texttt{melanoma} & $\mathbf{81.60}$ & $58.30$ & $65.00$ & $76.00$ \\
\texttt{mgus} & $\mathbf{88.40}$ & $46.00$ & $80.20$ & $86.10$ \\
\texttt{nki70} & $96.30$ & $28.50$ & $\mathbf{98.80}$ & $95.10$ \\
\texttt{pbc} & $\mathbf{74.40}$ & $47.70$ & $52.50$ & $64.70$ \\
\texttt{pbc3} & $\mathbf{91.20}$ & $32.80$ & $71.40$ & $89.40$ \\
\texttt{prostate} & $\mathbf{92.20}$ & $58.50$ & $82.20$ & $91.90$ \\
\texttt{stagec} & $\mathbf{86.70}$ & $44.10$ & $53.50$ & $78.50$ \\
\texttt{uis} & $\mathbf{87.90}$ & $62.10$ & $67.70$ & $85.10$ \\
\bottomrule
\end{tabularx}
\end{table}

\begin{table*}[!t]
   \centering
   \setlength{\tabcolsep}{3pt}
   \caption{Per-dataset downstream utility under TSTR: random baseline versus FoGS on C-index, IBS, and NNDR (oriented so that a positive value denotes an improvement over the baseline for both metrics). Reported values are $\times 100$. Bottom rows give the aggregates and the one-sided Wilcoxon $p$.}
   \label{tab:results_unfiltered}
   \small
   \begin{tabularx}{\textwidth}{@{}l*{9}{Y}@{}}
       \toprule
       & \multicolumn{3}{c}{C-index $\uparrow$}
       & \multicolumn{3}{c}{IBS $\downarrow$}
       & \multicolumn{3}{c}{NNDR $\uparrow$}\\
       \cmidrule(lr){2-4} \cmidrule(lr){5-7} \cmidrule(lr){8-10}
       \textbf{Dataset}
       & \textbf{Random} & \textbf{FoGS} & $\boldsymbol{\Delta}$
       & \textbf{Random} & \textbf{FoGS} & $\boldsymbol{\Delta}$
       & \textbf{Random} & \textbf{FoGS} & $\boldsymbol{\Delta}$ \\
       \midrule
       \texttt{gbsg}           & $68.85$ & $\mathbf{69.72}$ & \cellcolor{pgreen!25}$+0.87$ & $10.85$ & $\mathbf{9.99}$  & \cellcolor{pgreen!25}$+0.86$ & $\mathbf{75.01}$ & $74.30$          & \cellcolor{pred!25}$-0.71$ \\
       \texttt{metabric}       & $63.56$ & $\mathbf{64.28}$ & \cellcolor{pgreen!25}$+0.71$ & $10.19$ & $\mathbf{9.91}$  & \cellcolor{pgreen!25}$+0.28$ & $74.61$          & $\mathbf{80.79}$ & \cellcolor{pgreen!25}$+6.19$ \\
       \texttt{whas}           & $\mathbf{74.87}$ & $73.28$          & \cellcolor{pred!25}$-1.60$  & $\mathbf{10.67}$ & $11.08$          & \cellcolor{pred!25}$-0.41$  & $74.85$          & $\mathbf{85.20}$ & \cellcolor{pgreen!25}$+10.34$ \\
       \texttt{aids}           & $\mathbf{65.16}$ & $63.23$          & \cellcolor{pred!25}$-1.94$  & $\mathbf{4.45}$  & $4.75$           & \cellcolor{pred!25}$-0.30$  & $80.27$          & $\mathbf{82.49}$ & \cellcolor{pgreen!25}$+2.22$ \\
       \texttt{breast\_cancer} & $66.67$ & $\mathbf{71.65}$ & \cellcolor{pgreen!25}$+4.98$ & $9.94$  & $\mathbf{7.87}$  & \cellcolor{pgreen!25}$+2.06$ & $\mathbf{79.18}$ & $71.41$          & \cellcolor{pred!25}$-7.77$ \\
       \texttt{cost}           & $67.21$ & $\mathbf{67.45}$ & \cellcolor{pgreen!25}$+0.24$ & $16.74$ & $\mathbf{16.45}$ & \cellcolor{pgreen!25}$+0.29$ & $\mathbf{69.43}$ & $57.44$          & \cellcolor{pred!25}$-11.99$ \\
       \texttt{d\_oropha\_rec} & $53.25$ & $\mathbf{56.49}$ & \cellcolor{pgreen!25}$+3.25$ & $16.87$ & $\mathbf{16.40}$ & \cellcolor{pgreen!25}$+0.46$ & $56.36$          & $\mathbf{64.31}$ & \cellcolor{pgreen!25}$+7.95$ \\
       \texttt{hepatocellular} & $67.74$ & $\mathbf{76.91}$ & \cellcolor{pgreen!25}$+9.17$ & $14.81$ & $\mathbf{11.25}$ & \cellcolor{pgreen!25}$+3.56$ & $75.99$          & $\mathbf{84.13}$ & \cellcolor{pgreen!25}$+8.15$ \\
       \texttt{melanoma}       & $81.89$ & $\mathbf{82.94}$ & \cellcolor{pgreen!25}$+1.05$ & $9.84$  & $\mathbf{7.41}$  & \cellcolor{pgreen!25}$+2.43$ & $70.72$          & $\mathbf{71.02}$ & \cellcolor{pgreen!25}$+0.30$ \\
       \texttt{mgus}           & $69.98$ & $\mathbf{71.31}$ & \cellcolor{pgreen!25}$+1.33$ & $\mathbf{11.15}$ & $11.24$          & \cellcolor{pred!25}$-0.08$  & $75.27$          & $\mathbf{82.21}$ & \cellcolor{pgreen!25}$+6.93$ \\
       \texttt{nki70}          & $70.91$ & $\mathbf{74.55}$ & \cellcolor{pgreen!25}$+3.64$ & $12.63$ & $\mathbf{11.15}$ & \cellcolor{pgreen!25}$+1.48$ & $\mathbf{79.64}$ & $70.57$          & \cellcolor{pred!25}$-9.08$ \\
       \texttt{pbc}            & $74.36$ & $\mathbf{84.43}$ & \cellcolor{pgreen!25}$+10.07$ & $9.33$  & $\mathbf{6.96}$  & \cellcolor{pgreen!25}$+2.37$ & $58.64$          & $\mathbf{59.98}$ & \cellcolor{pgreen!25}$+1.34$ \\
       \texttt{pbc3}           & $77.87$ & $\mathbf{80.98}$ & \cellcolor{pgreen!25}$+3.11$ & $7.42$  & $\mathbf{6.33}$  & \cellcolor{pgreen!25}$+1.10$ & $71.22$          & $\mathbf{79.22}$ & \cellcolor{pgreen!25}$+8.00$ \\
       \texttt{prostate}       & $57.79$ & $\mathbf{58.84}$ & \cellcolor{pgreen!25}$+1.05$ & $\mathbf{16.84}$ & $18.45$          & \cellcolor{pred!25}$-1.62$  & $80.97$          & $\mathbf{90.50}$ & \cellcolor{pgreen!25}$+9.53$ \\
       \texttt{stagec}         & $63.04$ & $\mathbf{70.16}$ & \cellcolor{pgreen!25}$+7.11$ & $14.32$ & $\mathbf{10.27}$ & \cellcolor{pgreen!25}$+4.06$ & $65.47$          & $\mathbf{70.52}$ & \cellcolor{pgreen!25}$+5.05$ \\
       \texttt{uis}            & $\mathbf{62.21}$ & $61.87$          & \cellcolor{pred!25}$-0.34$  & $\mathbf{9.81}$  & $9.93$           & \cellcolor{pred!25}$-0.12$  & $75.64$          & $\mathbf{84.36}$ & \cellcolor{pgreen!25}$+8.72$ \\
       \midrule
       Mean                    & -- & -- & $\boldsymbol{+2.67}$ & -- & -- & $\boldsymbol{+1.03}$ & -- & -- & $\boldsymbol{+2.82}$ \\
       Median                  & -- & -- & $\boldsymbol{+1.19}$ & -- & -- & $\boldsymbol{+0.66}$ & -- & -- & $\boldsymbol{+5.62}$ \\
       $p$ (Wilcoxon)         & -- & -- & $0.007$              & -- & -- & $0.018$              & -- & -- & $0.087$\\
       Wins                    & -- & -- & $13/16$              & -- & -- & $11/16$              & -- & -- & $12/16$ \\
       \bottomrule
   \end{tabularx}
\end{table*}

\section{UMAP Projections of Real and Synthetic Cohorts}
\label{app:umap}

\begin{figure*}[p]
    \centering
    \includegraphics[width=\textwidth]{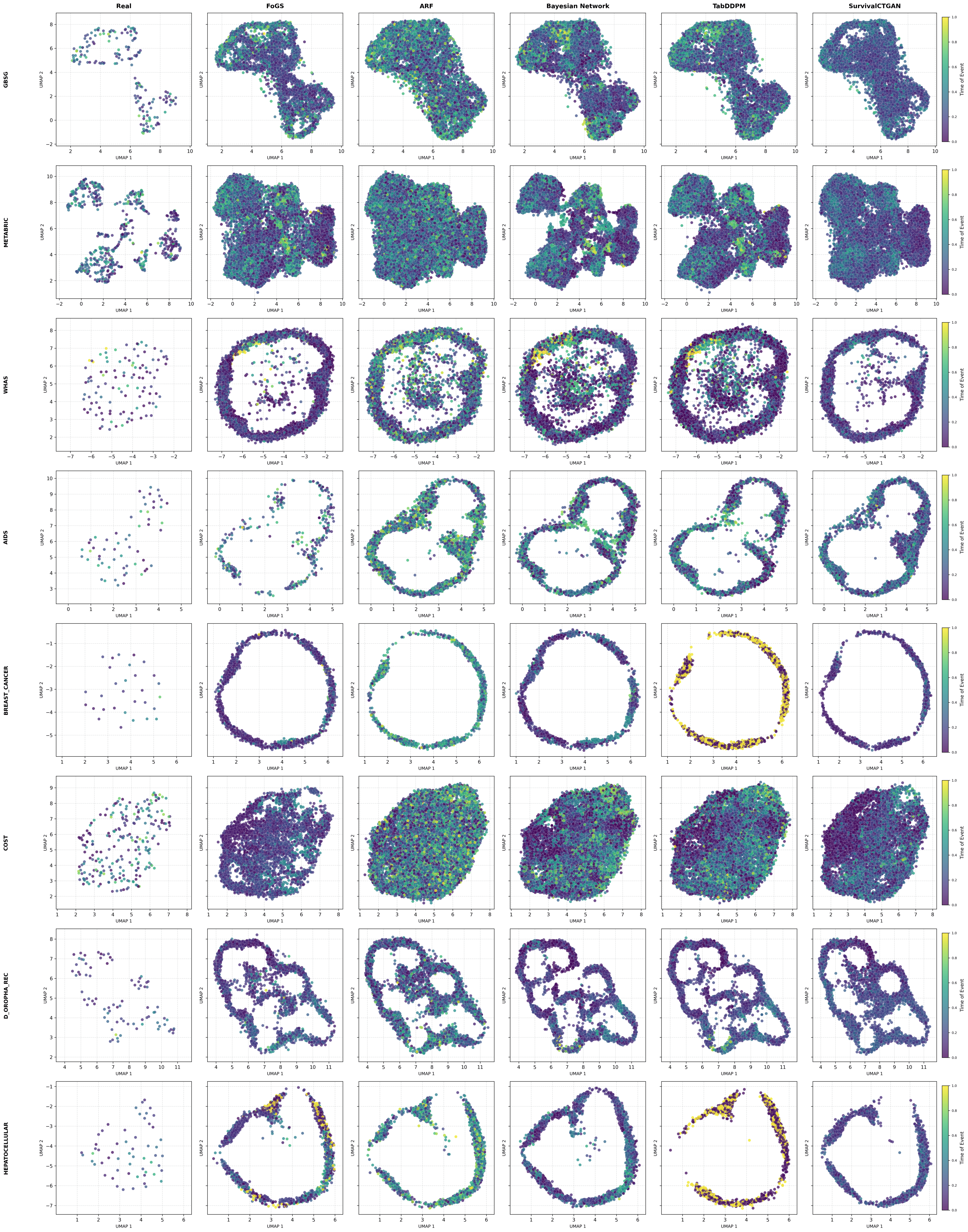}
     \vspace{-10pt}
    \caption{UMAP projections of the real and synthetic cohorts (part 1 of 2): datasets \texttt{gbsg}, \texttt{metabric}, \texttt{whas}, \texttt{aids}, \texttt{breast\_cancer}, \texttt{cost}, \texttt{d\_oropha\_rec}, and \texttt{hepatocellular} (rows). Columns, left to right: the real training data, the FoGS-selected set $\mathcal{D}_s^{*}$, and the over-generated candidate pools $\mathcal{D}_g$ of ARF, Bayesian Network, TabDDPM, and SurvivalCTGAN. For each dataset a single UMAP embedding is fitted on the real event-bearing samples and shared across columns, so the axes are comparable within a row; only event-bearing samples are shown, and point colour encodes the observed time, normalized to $[0,1]$ and discretized into five equal-width bins. Datasets \texttt{melanoma}--\texttt{uis} are shown in Fig.~\ref{fig:umap-part2}.}
    \label{fig:umap-part1}
\end{figure*}

\begin{figure*}[p]
    \centering
    \includegraphics[width=\textwidth]{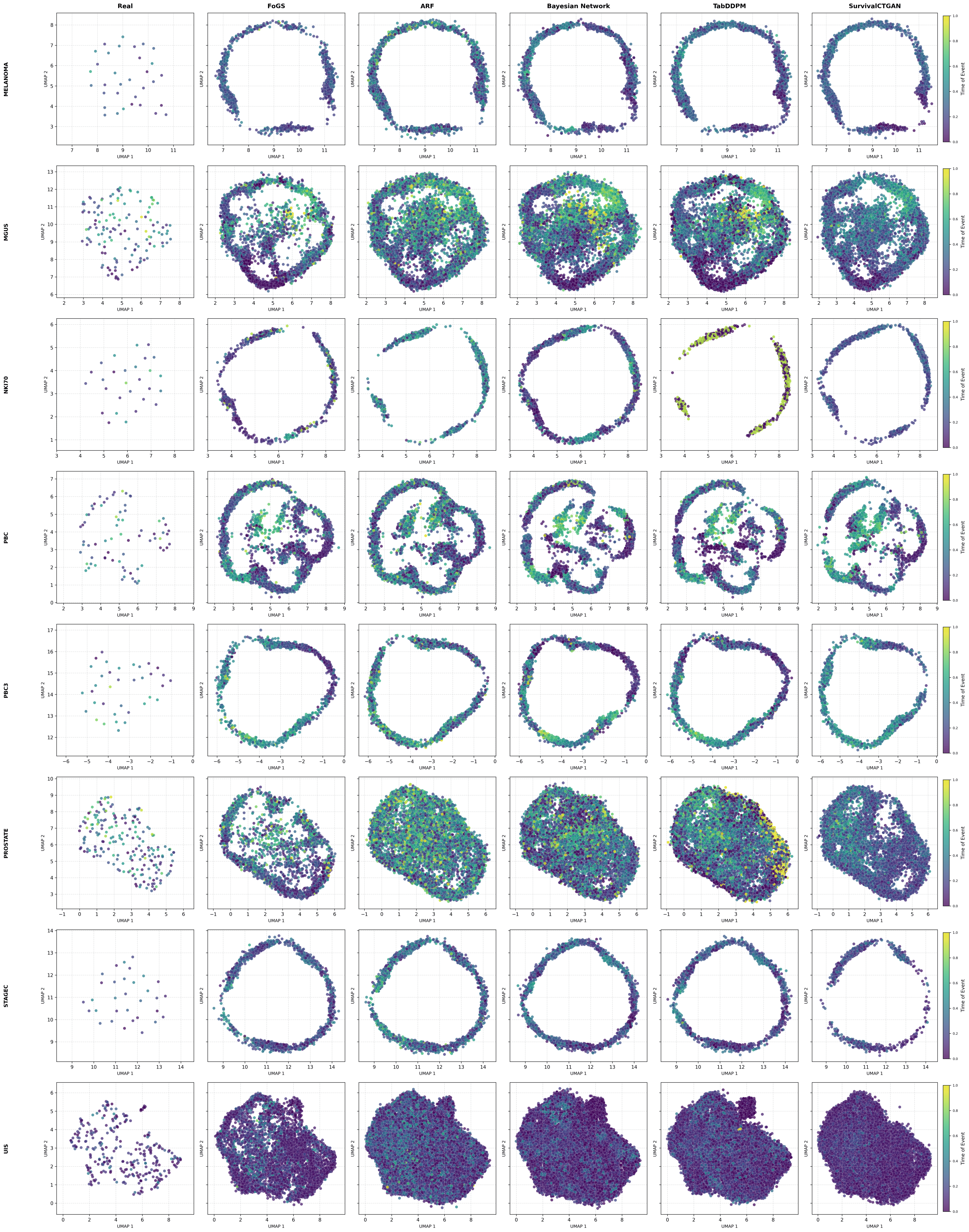}
    \vspace{-10pt}
    \caption{UMAP projections of the real and synthetic cohorts (part 2 of 2): datasets \texttt{melanoma}, \texttt{mgus}, \texttt{nki70}, \texttt{pbc}, \texttt{pbc3}, \texttt{prostate}, \texttt{stagec}, and \texttt{uis} (rows). Columns, embedding, and colour scale are as in Fig.~\ref{fig:umap-part1}.}
    \label{fig:umap-part2}
\end{figure*}

Figs.~\ref{fig:umap-part1} and~\ref{fig:umap-part2} provide a qualitative complement to the quantitative evaluation, projecting the real and synthetic cohorts of all 16 datasets into two dimensions with UMAP. For each dataset a single embedding is fitted on the real event-bearing samples ($\delta = 1$) and then applied unchanged to the event-bearing samples of every column; censored samples are not shown, so that the observed time coincides with the event time and can be used to colour the points. The embedding is therefore shared within a dataset, and the axes are directly comparable across its panels. From left to right, the columns show the real training data, the FoGS-selected set $\mathcal{D}_s^{*}$, and the over-generated candidate pools $\mathcal{D}_g$ ($N_g = \kappa N_r$) of the four single generators (ARF, Bayesian Network, TabDDPM, SurvivalCTGAN). The real cohorts sample the covariate space sparsely, whereas the candidate pools are dense, and the extent to which the four generators differ from one another varies across cohorts.

\end{document}